%% file: acl2016.tex
\documentclass[11pt]{article}
\usepackage{acl2016}
\usepackage{times}
\usepackage{url}
\usepackage{latexsym}
\usepackage{amsmath}
\usepackage{graphicx}
\usepackage{subfigure}

\usepackage{footnote}
\makesavenoteenv{tabular}
\makesavenoteenv{table}

\aclfinalcopy 
\def\aclpaperid{592} 

\title{Multi-Task Cross-Lingual Sequence Tagging from Scratch}

\author{
  Zhilin Yang ~~ Ruslan Salakhutdinov ~~ William Cohen \\
  School of Computer Science \\
  Carnegie Mellon University \\
  {\tt \{zhiliny,rsalakhu,wcohen\}@cs.cmu.edu}
}

\date{}

\begin{document}
\maketitle

\input{abstract}

\input{intro}

\input{related}

\input{model}

\input{exp}

\input{conc}

\section*{Acknowledgements}

This work was funded by the NSF under grant IIS-1250956.

\bibliography{acl2016}
\bibliographystyle{acl2016}

\end{document}

%% file: abstract.tex
\begin{abstract}
We present a deep hierarchical recurrent neural network for sequence
tagging. Given a sequence of words, our model employs deep gated
recurrent units on both character and word levels to encode morphology
and context information, and applies a conditional random field layer
to predict the tags. Our model is task independent, language
independent, and feature engineering free. We further extend our model
to multi-task and cross-lingual joint training by sharing the
architecture and parameters. Our model achieves state-of-the-art
results in multiple languages on several benchmark tasks including POS
tagging, chunking, and NER. We also demonstrate that multi-task and
cross-lingual joint training can improve the performance in various
cases.
\end{abstract}

%% file: intro.tex
\section{Introduction}

Sequence tagging is a fundamental problem in natural language
processing which has many wide applications, including part-of-speech
(POS) tagging, chunking, and named entity recognition (NER). Given a
sequence of words, sequence tagging aims to predict a linguistic tag
for each word such as the POS tag. Recently progress has been made on
neural sequence-tagging models which make only minimal assumptions
about the language, task, and feature set \cite{collobert2011natural}


This paper explores an important potential advantage of these
task-independent, language-independent and feature-engineering free
models: their ability to be jointly trained on multiple tasks.  In
particular, we explore two types of joint training. In
\textit{multi-task joint training}, a model is jointly trained to
perform multiple sequence-tagging tasks in the same language---e.g., POS
tagging and NER for English.  In \textit{cross-lingual joint
  training}, a model is trained to perform the same task in multiple
languages---e.g., NER in English and Spanish.

Multi-task joint training can exploit the fact that different sequence
tagging tasks in one language share language-specific
regularities. For example, models of English POS tagging and English
NER might benefit from using similar underlying representations for
words, and in past work, certain sequence-tagging tasks have
benefitted by leveraging the underlying similarity of related tasks
\cite{ando2005framework}. Currently, however, the best results on
specific sequence-tagging tasks are usually achieved by approaches
that target only one specific task, either POS tagging
\cite{sogaard2011semisupervised,toutanova2003feature}, chunking
\cite{shen2005voting}, or NER
\cite{luo2015joint,passos2014lexicon}. Such approaches employ
separate model development for each individual task, which makes joint
training difficult.  In other work, some recent neural approaches have
been proposed to address multiple sequence tagging problems in a
unified framework \cite{huang2015bidirectional}. Though gains have been
shown using multi-task joint training, the
prior models that benefit from multi-task joint training did not achieve
state-of-the-art performance \cite{collobert2011natural}; thus the
question of whether joint training can improve over strong baseline
methods is still unresolved.



Cross-lingual joint training typically uses word alignments or
parallel corpora to improve the performance on different languages
\cite{kiros2014multiplicative,gouws2014bilbowa}. However, many
successful approaches in sequence tagging rely heavily on feature
engineering to handcraft language-dependent features, such as
character-level morphological features and word-level N-gram patterns
\cite{huang2015bidirectional,toutanova2003feature,sun2008modeling},
making it difficult to share latent representations between different
languages. Some multilingual taggers that do not rely on feature
engineering have also been presented
\cite{lample2016neural,dos2015boosting}, but while these methods are
language-independent, they do not study the effect of cross-lingual
joint training.



In this work, we focus on developing a general model that can be applied in both multi-task and cross-lingual settings by learning from scratch, i.e., without feature engineering or pipelines. Given a sequence of words, our model employs deep gated recurrent units on both character and word levels, and applies a conditional random field layer to make the structured prediction. On the character level, the gated recurrent units capture the morphological information; on the word level, the gated recurrent units learn N-gram patterns and word semantics.

Our model can handle both multi-task and cross-lingual joint training
in a unified manner by simply sharing the network architecture and
model parameters between tasks and languages. For multi-task joint
training, we share both character and word level parameters between
tasks to learn language-specific regularities. For cross-lingual joint
training, we share the character-level parameters to capture the
morphological similarity between languages without use of parallel
corpora or word alignments.


We evaluate our model on five datasets of different tasks and
languages, including POS tagging, chunking and NER in English; and NER
in Dutch and Spanish.  We achieve state-of-the-art results on several
standard benchmarks: CoNLL 2000 chunking (95.41\%), CoNLL 2002 Dutch
NER (85.19\%), CoNLL 2003 Spanish NER (85.77\%), and CoNLL 2003
English NER (91.20\%). We also achieve very competitive results on
Penn Treebank POS tagging (97.55\%, the second best result in the literature).
Finally, we
conduct experiments to systematically explore the effectiveness of
multi-task and cross-lingual joint training on several tasks.

%% file: related.tex
\section{Related Work}

Ando and Zhang \shortcite{ando2005framework} proposed a multi-task joint training framework that shares structural parameters among multiple tasks, and improved the performance on various tasks including NER. Collobert et al. \shortcite{collobert2011natural} presented a task independent convolutional network and employed multi-task joint training to improve the performance of chunking. However, there is still a gap between these multi-task approaches and the state-of-the-art results on individual tasks. Furthermore, it is unclear whether these approaches can be effective in a cross-lingual setting.

Multilingual resources were extensively used for
cross-lingual sequence tagging through various ways, such as
cross-lingual feature extraction \cite{darwish2013named}, text
categorization \cite{virga2003transliteration}, and Bayesian parallel
data prediction \cite{snyder2008unsupervised}. Parallel corpora and
word alignments are also used for training cross-lingual distributed
word representations
\cite{kiros2014multiplicative,gouws2014bilbowa,zhou2015learning}. Unlike
these approaches, our method mainly focuses on using morphological
similarity for cross-lingual joint training.

Several neural architectures based on recurrent networks were proposed
for sequence tagging. Huang et al. \shortcite{huang2015bidirectional}
used word-level Long Short-Term Memory (LSTM) units based on
handcrafted features; dos Santos et al. \shortcite{dos2015boosting}
employed convolutional layers on both character and word levels; Chiu
and Nichols \shortcite{chiu2015named} applied convolutional layers on
the character level and LSTM units on the word level; Gillick et al. \shortcite{gillick2015multilingual} employed a sequence-to-sequence LSTM with a novel tagging scheme. We show that our
architecture gives better performance experimentally than these
approaches in Section \ref{sec:exp}.

Most similar to our work is the recent approach independently
developed by Lample et al. \shortcite{lample2016neural} (published two
weeks before our submission), which employs LSTM on both character and
word levels. However, there are several crucial differences. First, we
study cross-lingual joint training and show improvement over their
approach in various cases. Second, while they mainly focus on NER, we
generalize our model to other sequence tagging tasks, and also
demonstrate the effectiveness of multi-task joint training. There are
also differences in the technical aspect, such as the
cost-sensitive loss function and gated recurrent units used in our
work.

%% file: model.tex
\section{Model}

\begin{figure*}[t]
  \centering
  \vskip 0.1in
    \includegraphics[width=0.85\textwidth]{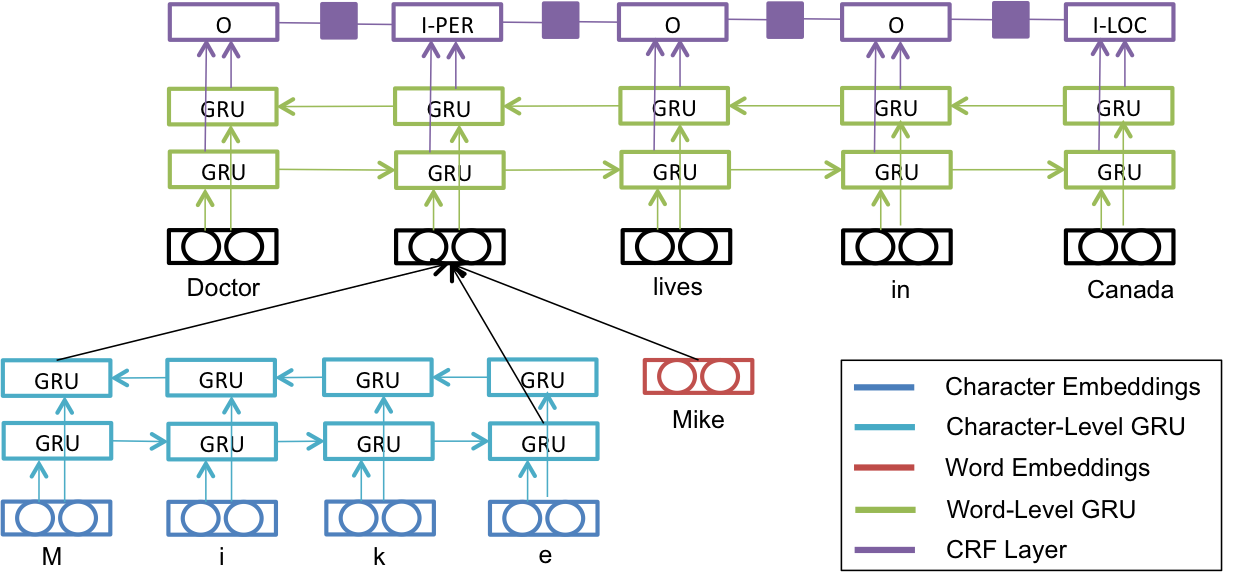}
  \caption{ \small The architecture of our hierarchical GRU network with CRF, when $L^c = L^w = 1$ (only one layer for word-level and character-level GRUs respectively). We only display the character-level GRU for the word \textit{Mike} and omit others.}
  \label{fig:arc}
\end{figure*}

In this section, we present our model for sequence tagging based on deep hierarchical gated recurrent units and conditional random fields. Our recurrent networks are \textit{hierarchical} since we have multiple layers on both word and character levels in a hierarchy.

\subsection{Gated Recurrent Unit}

A gated recurrent unit (GRU) network is a type of recurrent neural networks first introduced for machine translation \cite{cho2014properties}. A recurrent network can be represented as a sequence of units, corresponding to the input sequence $(\mathbf{x}_1, \mathbf{x}_2, \cdots, \mathbf{x}_T)$, which can be either a word sequence in a sentence or a character sequence in a word. The unit at position $t$ takes $\mathbf{x}_t$ and the previous hidden state $\mathbf{h}_{t - 1}$ as input, and outputs the current hidden state $\mathbf{h}_t$. The model parameters are shared between different units in the sequence.

A gated recurrent unit at position $t$ has two gates, an update gate $\mathbf{z}_t$ and a reset gate $\mathbf{r}_t$. More specifically, each gated recurrent unit can be expressed as follows
\begin{eqnarray*}
\mathbf{r}_t &=& \sigma(W_{rx} \mathbf{x}_t + W_{rh} \mathbf{h}_{t - 1}) \\
\mathbf{z}_t &=& \sigma(W_{zx} \mathbf{x}_t + W_{zh} \mathbf{h}_{t - 1}) \\
\tilde{\mathbf{h}}_t &=& \tanh(W_{hx} \mathbf{x}_t + W_{hh} (\mathbf{r}_t \odot \mathbf{h}_{t - 1})) \\
\mathbf{h}_t &=& \mathbf{z}_t \odot \mathbf{h}_{t- 1} + (1 - \mathbf{z}_t) \odot \tilde{\mathbf{h}}_t,
\end{eqnarray*}
where $W$'s are model parameters of each unit, $\tilde{\mathbf{h}}_t$ is a candidate hidden state that is used to compute $\mathbf{h}_t$, $\sigma$ is an element-wise sigmoid logistic function defined as $\sigma(\mathbf{x}) = 1 / (1 + e^{- \mathbf{x}})$, and $\odot$ denotes element-wise multiplication of two vectors. Intuitively, the update gate $\mathbf{z}_t$ controls how much the unit updates its hidden state, and the reset gate $\mathbf{r}_t$ determines how much information from the previous hidden state needs to be reset.

Since a recurrent neural network only models the information flow in one direction, it is usually helpful to use an additional recurrent network that goes in the reverse direction. More specifically, we use bidirectional gated recurrent units, where given a sequence of length $T$, we have one GRU going from $1$ to $T$ and the other from $T$ to $1$. Let $\overrightarrow{\mathbf{h}}_t$ and $\overleftarrow{\mathbf{h}}_t$ denote the hidden states at position $t$ of the forward and backward GRUs respectively. We concatenate the two hidden states to form the final hidden state $\mathbf{h}_t = [\overrightarrow{\mathbf{h}}_t, \overleftarrow{\mathbf{h}}_t]$.

We stack multiple recurrent layers together to form a deep recurrent network \cite{sutskever2014sequence}. Each layer learns a more effective representation taking the hidden states of the previous layer as input. Let $\mathbf{h}_{l, t}$ denote the hidden state at position $t$ in layer $l$. The forward GRU at position~$t$ in layer $l$ computes $\overrightarrow{\mathbf{h}}_{l, t}$ using $\overrightarrow{\mathbf{h}}_{l, t - 1}$ and $\mathbf{h}_{l - 1, t}$ as input, and the backward GRU performs similar operations but in a reverse direction.

\subsection{Hierarchical GRU}

Our model employs a hierarchical GRU that encodes both word-level and character-level sequential information.

The input of our model is a sequence of words $(\mathbf{x}_1, \mathbf{x}_2, \cdots, \mathbf{x}_T)$ of length $T$, where $\mathbf{x}_t$ is a one-of-$K$ embedding of the $t$-th word. The word at each position $t$ also has a character-level representation, denoted as a sequence of length $S_t$, $(\mathbf{c}_{t, 1}, \mathbf{c}_{t, 2}, \cdots, \mathbf{c}_{t, S_t})$ where $\mathbf{c}_{t, s}$ is the one-of-$K$ embedding of the $s$-th character in the $t$-th word.

\subsubsection{Character-Level GRU}

Given a word, we first employ a deep bidirectional GRU to learn useful morphological representation from the character sequence of the word. Suppose the character-level GRU has $L^c$ layers, we then obtain forward and backward hidden states $\overleftarrow{\mathbf{h}}_{L^c, s}$ and $\overrightarrow{\mathbf{h}}_{L^c, s}$ at each position $s$ in the character sequence. Since recurrent networks usually tend to memorize more short-term patterns, we concatenate the first hidden state of the backward GRU and the last hidden state of the forward GRU to encode character-level morphology in both prefixes and suffixes. We further concatenate the character-level representation with the one-of-$K$ word embedding $\mathbf{x}_t$ to form the final representation $\mathbf{h}^w_t$ for the $t$-th word. More specifically, we have
\[
\mathbf{h}^w_t = [\overrightarrow{\mathbf{h}}_{L^c, S_t}, \overleftarrow{\mathbf{h}}_{L^c, 1}, \mathbf{x}_t],
\]
where $\mathbf{h}^w_t$ is a representation of the $t$-th word, which encodes both character-level morphology and word-level semantics, as shown in Figure~\ref{fig:arc}.

\subsubsection{Word-Level GRU}

The character-level GRU outputs a sequence of word representations $\mathbf{h}^w = (\mathbf{h}^w_1, \mathbf{h}^w_2, \cdots, \mathbf{h}^w_T)$. We employ a word-level deep bidirectional GRU with $L^w$ layers on top of these word representations. The word-level GRU takes the sequence $\mathbf{h}^w$ as input, and computes a sequence of hidden states $\mathbf{h} = (\mathbf{h}_1, \mathbf{h}_2, \cdots, \mathbf{h}_T)$.

Different from the character-level GRU, the word-level GRU aims to extract the context information in the word sequence, such as N-gram patterns and neighbor word dependencies. 
Such information is usually encoded using handcrafted features. However, 
as we show in our experimental results, the word-level GRU can learn the relevant
information without being language-specific or task-specific.
The hidden states $\mathbf{h}$ output by the word-level GRU will be used as input features for the next layers.

\subsection{Conditional Random Field}

The goal of sequence tagging is to predict a sequence of tags $\mathbf{y} = (y_1, y_2, \cdots, y_T)$. To model the dependencies between tags in a sequence, we apply a conditional random field \cite{lafferty2001conditional} layer on top of the hidden states $\mathbf{h}$ output by the word-level GRU \cite{huang2015bidirectional}. Let $\mathcal{Y}(\mathbf{h})$ denote the space of tag sequences for $\mathbf{h}$. The conditional log probability of a tag sequence $\mathbf{y}$, given the hidden state sequence $\mathbf{h}$, can be written as
\begin{equation} \label{eq:cond}
\log p(\mathbf{y} | \mathbf{h}) = f(\mathbf{h}, \mathbf{y}) - \log \sum_{\mathbf{y}' \in \mathcal{Y}(\mathbf{h})} \exp f(\mathbf{h}, \mathbf{y}'),
\end{equation}
where $f$ is a function that assigns a score for each pair of $\mathbf{h}$ and $\mathbf{y}$.

To define the function $f(\mathbf{h}, \mathbf{y})$, for each position~$t$, we multiply the hidden state $\mathbf{h}^w_t$ with a parameter vector $\mathbf{w}_{y_t}$ that is indexed by the the tag $y_t$, to obtain the score for assigning $y_t$ at position~$t$. Since we also need to consider the correlation between tags, we impose first order dependency by adding a score $A_{y_{t - 1}, y_t}$ at position $t$, where $A$ is a parameter matrix defining the similarity scores between different tag pairs. Formally, the function $f$ can be written as
\[
f(\mathbf{h}, \mathbf{y}) = \sum_{t = 1}^T \mathbf{w}_{y_t}^T \mathbf{h}^w_t + \sum_{t = 1}^T A_{y_{t - 1}, y_t},
\]
where we set $y_0$ to be a \textsc{Start} token.

It is possible to directly maximize the conditional log likelihood based on Eq. (\ref{eq:cond}). However, this training objective is usually not optimal since each possible $\mathbf{y}'$ contributes equally to the objective function. Therefore, we add a cost function between $\mathbf{y}$ and $\mathbf{y}'$ based on the max-margin principle that high-cost tags $\mathbf{y}'$ should be penalized more heavily \cite{gimpel2010softmax}. More specifically, the objective function to maximize for each training instance $\mathbf{y}$ and $\mathbf{h}$ is written as
\begin{equation}
f(\mathbf{h}, \mathbf{y}) - \log \sum_{\mathbf{y}' \in \mathcal{Y}(\mathbf{h})} \exp (f(\mathbf{h}, \mathbf{y}') + \mbox{cost}(\mathbf{y}, \mathbf{y}')).
\label{eq:norm}
\end{equation}

In our work, the cost function is defined as the tag-wise Hamming loss between two tag sequences multiplied by a constant. The objective function on the training set is the sum of Eq. (\ref{eq:norm}) over all the training instances.
The full architecture of our model is illustrated in Figure~\ref{fig:arc}.

\subsection{Training}


We employ mini-batch AdaGrad \cite{duchi2011adaptive} to train our neural network in an end-to-end manner with backpropagation. Both the character embeddings and word embeddings are fine-tuned during training. We use dynamic programming to compute the normalizer of the CRF layer in Eq.~(\ref{eq:norm}). When making prediction, we again use dynamic programming in the CRF layer to decode the most probable tag sequence.

\section{Multi-Task and Cross-Lingual Joint Training} \label{sec:joint}

\begin{figure}[t]
\vskip 0.1in
\centering
\subfigure[Multi-Task Joint Training]{\label{fig:multi-task} \includegraphics[width = 0.9\columnwidth]{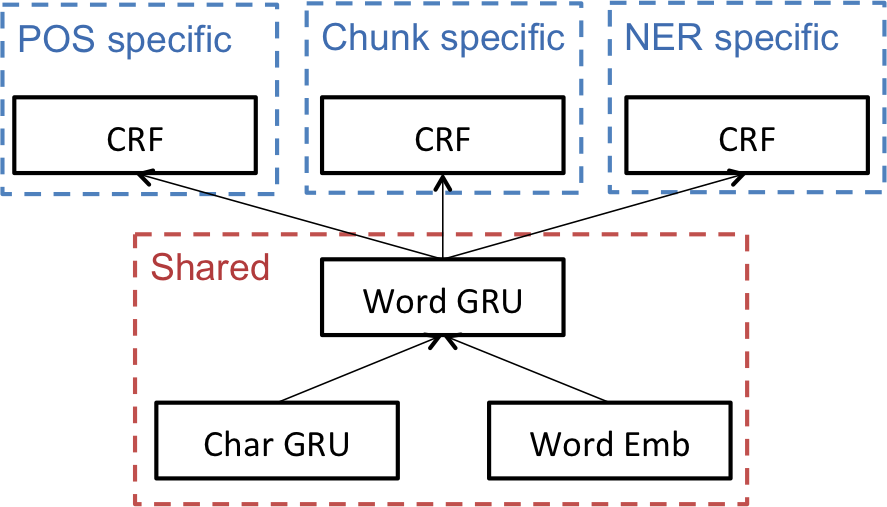}}
\vskip 0.1in
\subfigure[Cross-Lingual Joint Training]{\label{fig:cross-lingual} \includegraphics[width = 0.9\columnwidth]{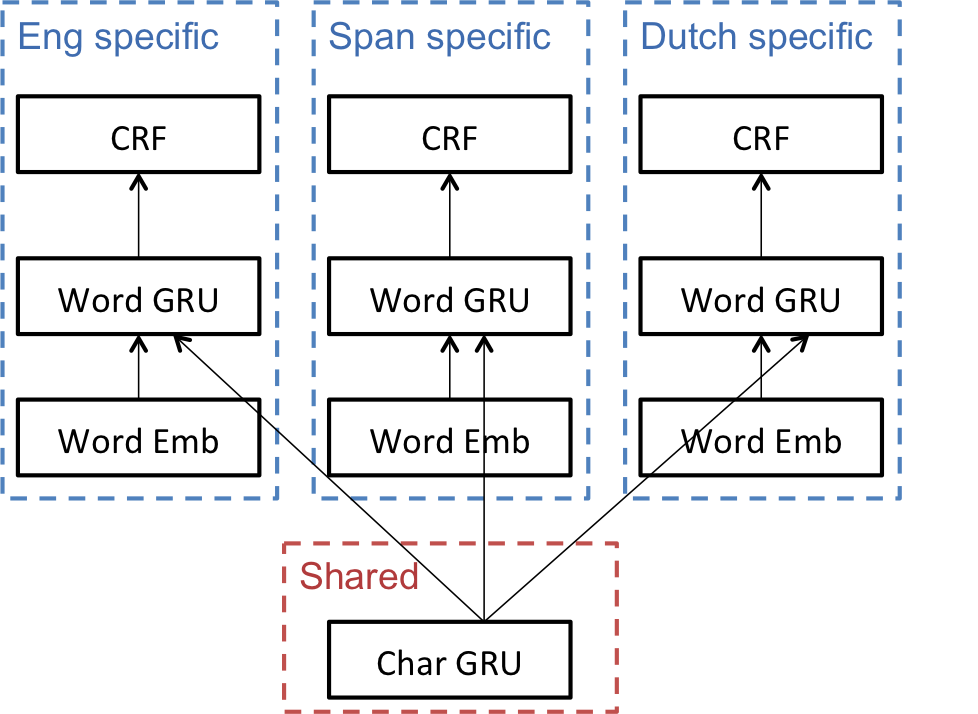}}
\caption{\small Network architectures for multi-task and cross-lingual joint training. Red boxes indicate shared architecture and parameters. Blue boxes are task/language specific components trained separately. \textit{Eng}, \textit{Span}, \textit{Char}, and \textit{Emb} refer to English, Spanish, Character and Embeddings.}
\label{fig:joint}
\end{figure} 

In this section we study joint training of multiple tasks and multiple languages. On one hand, different sequence tagging tasks in the same language share language-specific regularities. For example, POS tagging and NER in English should learn similar underlying representation since they are in the same language. On the other hand, some languages share character-level morphologies, such as English and Spanish. Therefore, it is desirable to leverage multi-task and cross-lingual joint training to boost model performance.

Since our model is generally applicable to different tasks in different languages, it can be naturally extended to multi-task and cross-lingual joint training. The basic idea is to share part of the architecture and parameters between tasks and languages, and to jointly train multiple objective functions with respect to different tasks and languages.

We now discuss the details of our joint training algorithm in the multi-task setting. Suppose we have $D$ tasks, with the training instances of each task being $(X_1, X_2, \cdots, X_D)$. Each task $d$ has a set of model parameters $W_d$, which is divided into two sets, \textit{task specific parameters} and \textit{shared parameters}, i.e.,
\[
W_d = W_{d, \mbox{spec}} \cup W_{\mbox{shared}},
\]
where shared parameters $W_{\mbox{shared}}$ are a set of parameters that are shared among the $D$ tasks, while task specific parameters $W_{d, \mbox{spec}}$ are the rest of the parameters that are trained for each task $d$ separately.

During joint training, we are optimizing the average over all objective functions of $D$ tasks. We iterate over each task $d$, sample a batch of training instances from $X_d$, and perform a gradient descent step to update model parameters $W_d$.
Similarly, we can derive a cross-lingual joint training algorithm by replacing $D$ tasks with $D$ languages.

The network architectures we employ for joint training are illustrated in Figure \ref{fig:joint}. For multi-task joint training, we share all the parameters below the CRF layer including word embeddings to learn language-specific regularities shared by the tasks. For cross-lingual joint training, we share the parameters of the character-level GRU to capture the morphological similarity between languages. Note that since we do not consider using parallel corpus in this work, we mainly focus on joint training between languages with similar morphology. We leave the study of cross-lingual joint training by sharing word semantics based on parallel corpora to future work.

%% file: exp.tex
\section{Experiments}
\label{sec:exp}

\begin{table*}[t]
\vspace{-0.1in}
\caption{Dataset Statistics}
\label{tab:stat}
\begin{center}
\begin{tabular}{lllrrr}
Benchmark & Task & Language & \# Training Tokens & \# Dev Tokens & \# Test Tokens
\\ \hline \\
PTB \shortcite{toutanova2003feature} & POS Tagging & English &  912,344 & 131,768 & 129,654 \\
CoNLL 2000 & Chunking & English & 211,727 & - & 47,377 \\
CoNLL 2003 & NER & English & 204,567 & 51,578 & 46,666 \\
CoNLL 2002 & NER & Dutch & 202,931 & 37,761 & 68,994 \\
CoNLL 2002 & NER & Spanish & 207,484 & 51,645 & 52,098
\end{tabular}
\end{center}
\vspace{-0.2in}
\end{table*}

In this section, we use several benchmark datasets for multiple tasks in multiple languages to evaluate our model as well as the joint training algorithm.

\subsection{Datasets and Settings}

We use the following benchmark datasets in our experiments: Penn Treebank (PTB) POS tagging, CoNLL 2000 chunking, CoNLL 2003 English NER, CoNLL 2002 Dutch NER and CoNLL 2002 Spanish NER. The statistics of the datasets are described in Table \ref{tab:stat}.

We construct the POS tagging dataset with the instructions described in Toutanova et al. \shortcite{toutanova2003feature}. Note that as a standard practice, the POS tags are extracted from the parsed trees.

For the task of CoNLL 2003 English NER, we follow previous works
\cite{collobert2011natural,huang2015bidirectional,chiu2015named} to
append one-hot gazetteer features to the input of the CRF layer for
fair comparison.\footnote{Although gazetteers are arguably a type of
  feature engineering, we note that unlike most feature engineering
  techniques they are straightforward to include in a model.  We use
  only the gazetteer file provided by the CoNLL 2003 shared task, and
  do not use gazetteers for any other tasks or languages described
  here.}

\begin{table}[t]
\caption{\small Comparison with state-of-the-art results on CoNLL 2003 English NER when trained with training set only. $^\dagger$ means using handcrafted features. $^\ddagger$ means being task-specific.}
\label{tab:ner-eng}
\begin{center}
\begin{tabular}{ll}
Model & F1 (\%)
\\ \hline \\
Chieu et al. \shortcite{chieu2002named}$^{\dagger\ddagger}$ & 88.31 \\
Florian et al. \shortcite{florian2003named}$^{\dagger\ddagger}$ & 88.76 \\
Ando and Zhang \shortcite{ando2005framework}$^{\dagger}$ & 89.31 \\
Lin and Wu \shortcite{lin2009phrase}$^{\dagger\ddagger}$ & 90.90 \\
Collobert et al. \shortcite{collobert2011natural} & 89.59 \\
Huang et al. \shortcite{huang2015bidirectional}$^\dagger$ & 90.10 \\ \hline
Ours & \textbf{90.94}
\end{tabular}
\end{center}
\vspace{-0.2in}
\end{table}

We set the hidden state dimensions to be 300 for the word-level GRU. We set the number of GRU layers to $L_c = L_w = 2$ (two layers for the word-level and character-level GRUs respectively). The learning rate is fixed at 0.01. We use the development set to tune the other hyperparameters of our model. Since the CoNLL 2000 chunking dataset does not have a development set, we hold out one fifth of the training set for parameter tuning.

\begin{table}[t!]
\caption{\small Comparison with state-of-the-art results on CoNLL 2003 English NER when trained with both training and dev sets. $^\dagger$ means using handcrafted features. $^\ddagger$ means being task-specific. $^*$ means not using gazetteer lists.}
\label{tab:ner-eng-dev}
\begin{center}
\begin{tabular}{ll}
Model & F1 (\%)
\\ \hline \\
Ratinov and Roth \shortcite{ratinov2009design}$^{\dagger\ddagger}$ & 90.80 \\
Passos et al. \shortcite{passos2014lexicon}$^{\dagger\ddagger}$ & 90.90 \\
Chiu and Nichols \shortcite{chiu2015named} & 90.77 \\
Luo et al. \shortcite{luo2015joint}$^{\dagger\ddagger}$ & \textbf{91.2} \\
Lample et al. \shortcite{lample2016neural}$^*$ & 90.94 \\ \hline
Ours & \textbf{91.20} \\
Ours $-$ no gazetteer$^*$ & 90.96 \\
Ours $-$ no char GRU & 88.00 \\
Ours $-$ no word embeddings & 77.20
\end{tabular}
\end{center}
\vspace{-0.2in}
\end{table}

We truncate all words whose character sequence length is longer than a threshold (17 for English, 35 for Dutch, and 20 for Spanish). We replace all numeric characters with ``0''.
We also use the BIOES (Begin, Inside, Outside, End, Single) tagging scheme \cite{ratinov2009design}.

\subsection{Pre-Trained Word Embeddings}

Since the training corpus for a sequence tagging task is relatively small, it is difficult to train randomly initialized word embeddings to accurately capture the word semantics. Therefore, we leverage word embeddings pre-trained on large-scale corpora. All the pre-trained embeddings we use are publicly available.

On the English datasets, following previous works that are based on neural networks \cite{collobert2011natural,huang2015bidirectional,chiu2015named}, we use the 50-dimensional SENNA embeddings\footnote{\url{http://ronan.collobert.com/senna/}} trained on Wikipedia.
For Spanish and Dutch, we use the 64-dimensional Polyglot embeddings\footnote{\url{https://sites.google.com/site/rmyeid/projects/polyglot}} \cite{al2013polyglot}, which are trained on Wikipedia articles of the corresponding languages.
We use pre-trained word embeddings as initialization, and fine-tune the embeddings during training.

\begin{table}[t]
\caption{\small Comparison with state-of-the-art results on CoNLL 2002 Dutch NER. $^\dagger$ means using handcrafted features. $^\ddagger$ means being task-specific.}
\label{tab:ner-ned}
\begin{center}
\begin{tabular}{ll}
Model & F1 (\%)
\\ \hline \\
Carreras et al. \shortcite{carreras2002named}$^{\dagger\ddagger}$ & 77.05 \\
Nothman et al. \shortcite{nothman2013learning}$^{\dagger\ddagger}$ & 78.6 \\
Gillick et al. \shortcite{gillick2015multilingual} & 82.84 \\
Lample et al. \shortcite{lample2016neural} & 81.74 \\ \hline
Ours & 85.00 \\
Ours $+$ joint training & \textbf{85.19} \\
Ours $-$ no char GRU & 77.76 \\
Ours $-$ no word embeddings & 67.36
\end{tabular}
\end{center}
\vspace{-0.2in}
\end{table}

\begin{table}[t]
\caption{\small Comparison with state-of-the-art results on CoNLL 2002 Spanish NER. $^\dagger$ means using handcrafted features. $^\ddagger$ means being task-specific.}
\label{tab:ner-esp}
\begin{center}
\begin{tabular}{ll}
Model & F1 (\%)
\\ \hline \\
Carreras et al. \shortcite{carreras2002named}$^{\dagger\ddagger}$ & 81.39 \\
dos Santos et al. \shortcite{dos2015boosting} & 82.21 \\
Gillick et al. \shortcite{gillick2015multilingual} & 82.95 \\
Lample et al. \shortcite{lample2016neural} & 85.75 \\ \hline
Ours & 84.69 \\
Ours $+$ joint training & \textbf{85.77} \\
Ours $-$ no char GRU & 83.03 \\
Ours $-$ no word embeddings & 73.34
\end{tabular}
\end{center}
\vspace{-0.2in}
\end{table}

\begin{figure}[t]
  \centering
    \includegraphics[width=0.5\textwidth]{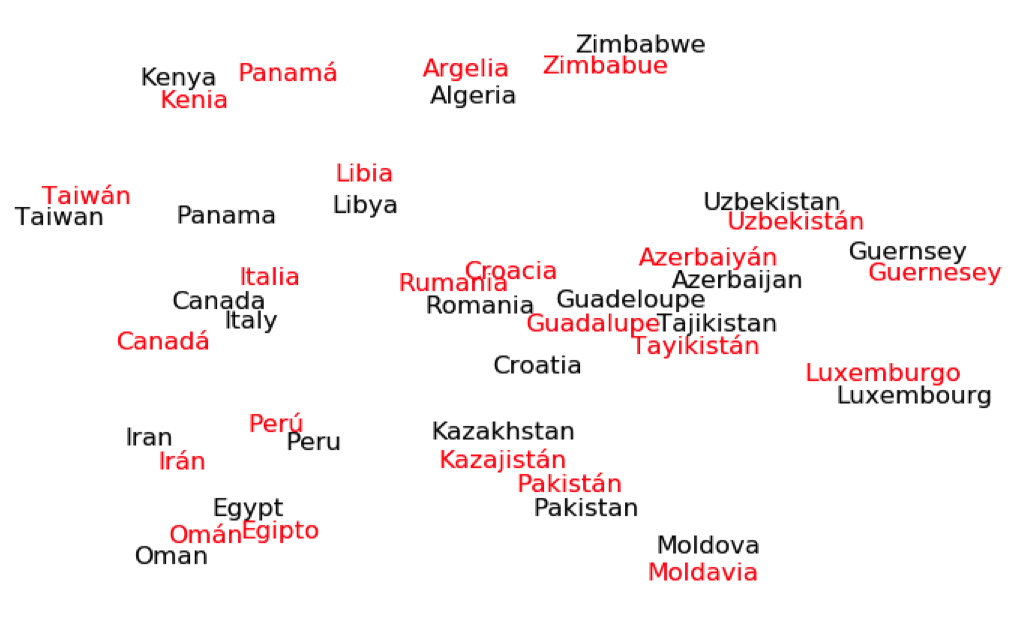}
\vskip -0.1in
  \caption{\small 2-dimensional t-SNE visualization of the character-level GRU output for country names in English and Spanish. Black words are English and red ones are Spanish. Note that all corresponding pairs are nearest neighbors in the original embedding space.}
  \label{fig:vis}
\end{figure}

\begin{table}[t]
\caption{\small Comparison with state-of-the-art results on CoNLL 2000 English chunking. $^\dagger$ means using handcrafted features. $^\ddagger$ means being task-specific.}
\label{tab:chunk}
\begin{center}
\begin{tabular}{ll}
Model & F1 (\%)
\\ \hline \\
Kudo and Matsumoto \shortcite{kudo2001chunking}$^{\dagger\ddagger}$ & 93.91 \\
Shen and Sarkar \shortcite{shen2005voting}$^{\dagger\ddagger}$ & 94.01\footnote{We note that this number is often mistakenly cited as 95.23, which is actually the score on base NP chunking rather than CoNLL 2000.} \\
Sun et al. \shortcite{sun2008modeling}$^{\dagger\ddagger}$ & 94.34 \\
Collobert et al. \shortcite{collobert2011natural} & 94.32 \\
Huang et al. \shortcite{huang2015bidirectional}$^\dagger$ & 94.46 \\ \hline
Ours & 94.66 \\
Ours $+$ joint training & \textbf{95.41} \\
Ours $-$ no char GRU & 94.44 \\
Ours $-$ no word embeddings & 88.13
\end{tabular}
\end{center}
\vspace{-0.2in}
\end{table}

\begin{table}[t]
\caption{\small Comparison with state-of-the-art results on PTB POS tagging. $^\dagger$ means using handcrafted features. $^\ddagger$ means being task-specific. $^*$ indicates our reimplementation (using SENNA embeddings).}
\label{tab:pos}
\begin{center}
\begin{tabular}{ll}
Model & Accuracy (\%)
\\ \hline \\
Toutanova et al. \shortcite{toutanova2003feature}$^{\dagger\ddagger}$ & 97.24 \\
Shen et al. \shortcite{shen2007guided}$^{\dagger\ddagger}$ & 97.33 \\
S{\o}gaard et al. \shortcite{sogaard2011semisupervised}$^{\dagger\ddagger}$ & 97.50 \\
Collobert et al. \shortcite{collobert2011natural} & 97.29 \\
Huang et al. \shortcite{huang2015bidirectional}$^\dagger$ & 97.55 \\
Ling et al. \shortcite{ling2015finding} & \textbf{97.78} \\
Ling et al. \shortcite{ling2015finding} (SENNA)$^*$ & 97.41 \\ \hline
Ours (SENNA) & {97.55} \\
Ours $-$ no char GRU & 96.69 \\
Ours $-$ no word embeddings & 95.43
\end{tabular}
\end{center}
\vspace{-0.2in}
\end{table}

\subsection{Performance}

In this section, we report the results of our model on the benchmark datasets and compare to the previously-reported state-of-the-art results.

For English NER, there are two evaluation methods used in the literature. Some models are trained with both the training and development set, while others are trained with the training set only. We report our results in both cases. In the first case, we tune the hyperparameters by training on the training set and testing on the development set.

Besides our standalone model, we experimented with multi-task and cross-lingual joint training as well, using the architecture described in Section \ref{sec:joint}. For multi-task joint training, we jointly train all tasks in English, including POS tagging, chunking and NER. For cross-lingual joint training, we jointly train NER in English, Dutch and Spanish.
We also remove the word embeddings and the character-level GRU respectively to analyze the contribution of different components.

The results are shown in Tables \ref{tab:ner-eng}, \ref{tab:ner-eng-dev}, \ref{tab:ner-ned}, \ref{tab:ner-esp}, \ref{tab:chunk} and \ref{tab:pos}. We achieve state-of-the-art results on English NER, Dutch NER, Spanish NER and English chunking. Our model outperforms the best previously-reported results on Dutch NER and English chunking by 2.35 points and 0.95 points respectively. We also achieve the second best result on English POS tagging, which is 0.23 points worse than the current state-of-the-art.

Joint training improves the performance on Spanish NER, Dutch NER and English chunking by 1.08 points, 0.19 points and 0.75 points respectively, and has no significant improvement on English POS tagging and English NER.

On POS tagging, the best result is 97.78\% reported by Ling et al. \shortcite{ling2015finding}. However, the embeddings they used are not publicly available. To demonstrate the effectiveness of our model, we slightly revise our model to reimplement their model
with the same parameter settings described in their original paper. We use SENNA embeddings to initialize the reimplemented model for fair comparison, and obtain an accuracy of 97.41\% that is 0.14 points worse than our result, which indicates that our model is more effective and the main difference lies in using different pre-trained embeddings.

By comparing the results without the character-level GRU and without word embeddings, we can observe that both components contribute to the final results. It is also clear that word embeddings have significantly more contribution than the character-level GRU, which indicates that our model largely depends on \textit{memorizing} the word semantics. Character-level morphology, on the other hand, has relatively smaller but still critical contribution.

\subsection{Joint Training}

In this section, we analyze the effectiveness of multi-task and
cross-lingual joint training in more detail. In order to explore
possible gains in performance of joint training for resource-poor
languages or tasks, we consider joint training of various task pairs
and language pairs where different-sized subsets of the actual labeled
corpora are made available. Given a pair of tasks of languages, we jointly train one task with full labels and the other with partial labels. In particular, we introduce a labeling
rate $r$, and sample a fraction $r$ of the sentences in the training
set, discarding the rest. Evaluation is based on the partially-labeled task. The results are reported in Table
\ref{tab:joint}.

\begin{table}[t]
\caption{\small Multi-task and cross-lingual joint training. We compare the results obtained by a standalone model and joint training with another task or language. The number following a task is the labeling rate (0.1 or 0.3). \textit{Eng} and \textit{NER} both refer to English NER, \textit{Span} means Spanish. In the column titles, \textit{Task} is the target task, \textit{J. Task} is the jointly-trained task with full labels, \textit{Sep.} is the F1/Accuracy of the target task trained separately, \textit{Joint} is the F1/Accuracy of the target task with joint training, and \textit{Delta} is the improvement.}
\label{tab:joint}
\begin{center}
\begin{tabular}{lllll}
Task & J. Task & Sep. & Joint & Delta
\\ \hline \\
Span 0.1 & Eng & 74.53 & 76.52 & $+$1.99 \\
Span 0.3 & Eng & 80.81 & 80.20 & $+$0.61 \\
Eng 0.1 & Span & 86.21 & 86.51 & $+$0.30 \\
Eng 0.3 & Span & 88.54 & 88.79 & $+$0.25 \\ \hline
POS 0.1 & NER & 96.59 & 96.79 & $+$0.20 \\
POS 0.3 & NER & 97.03 & 97.14 & $+$0.11 \\
NER 0.1 & POS & 86.21 & 87.02 & $+$0.81 \\
NER 0.3 & POS & 88.54 & 89.16 & $+$0.62 \\
Chunk 0.1 & NER & 90.65 & 91.16 & $+$0.51 \\
Chunk 0.3 & NER & 92.51 & 92.87 & $+$0.36 \\
\end{tabular}
\end{center}
\vspace{-0.2in}
\end{table}

We observe that the performance of a specific task with relatively lower labeling rates (0.1 and 0.3) can usually benefit from other tasks with full labels through multi-task or cross-lingual joint training. The performance gain can be up to 1.99 points when the labeling rate of the target task is~0.1.
The improvement with 0.1 labeling rate is on average 0.37 points larger than with 0.3 labeling rate, which indicates that the improvement of joint training is more significant when the target task has less labeled data.

We also use t-SNE \cite{van2008visualizing} to obtain a 2-dimensional visualization of the character-level GRU output for the country names in English and Spanish, shown in Figure~\ref{fig:vis}. We can clearly see that our model captures the morphological similarity between two languages through joint training, since all corresponding pairs are nearest neighbors in the original embedding space.

%% file: conc.tex
\section{Conclusion}

We presented a new model for sequence tagging based on gated recurrent units and conditional random fields. We explored multi-task and cross-lingual joint training through sharing part of the network architecture and model parameters. We achieved state-of-the-art results on various tasks including POS tagging, chunking, and NER, in multiple languages. We also demonstrated that joint training can improve model performance in various cases.

In this work, we mainly focus on leveraging morphological similarities for cross-lingual joint training. In the future, an important problem will be joint training based on cross-lingual word semantics with the help of parallel data. Furthermore, it will be interesting to apply our joint training approach to low-resource tasks and languages.